\documentclass[twocolumn]{IEEEtran}
\usepackage{lscape}
\usepackage{booktabs}
\usepackage{authblk}
\usepackage{url}
\usepackage{silence}
\usepackage[pdftex]{graphicx}
\usepackage{epstopdf}
\usepackage{pifont}
\usepackage{multirow}
\usepackage{float}
\usepackage{caption}
\usepackage[table,xcdraw]{xcolor}
\usepackage{array}
\usepackage{subcaption}
\usepackage{verse}
\usepackage{longtable}
\usepackage{supertabular}
\usepackage{cite}
\usepackage{color}
\usepackage{amsmath}
\usepackage{lettrine}
\usepackage{hyperref}
\usepackage{multirow}
\usepackage{multicol}
\usepackage{dirtytalk}
\usepackage{color}

\usepackage[colorinlistoftodos,prependcaption,textsize=tiny]{todonotes}

\hypersetup{
    colorlinks=false,
    linkcolor=blue,
    filecolor=magenta,
    urlcolor=cyan,
}

\begin{document}

\title{Visual Sentiment Analysis from Disaster Images in Social Media}


\author{
    \IEEEauthorblockN{Syed Zohaib\IEEEauthorrefmark{1}, Kashif Ahmad\IEEEauthorrefmark{2}, Steven Hicks\IEEEauthorrefmark{3},
    P{\aa}l Halvorsen\IEEEauthorrefmark{3},
    Ala Al-Fuqaha\IEEEauthorrefmark{2},
    Nicola Conci\IEEEauthorrefmark{1},
    Michael Riegler\IEEEauthorrefmark{3}}
    \IEEEauthorblockA{\IEEEauthorrefmark{1} University of Trento, Italy
    \\\{zohaib.hassan, nicola.conci\}@unitn.it} \\
    \IEEEauthorblockA{\IEEEauthorrefmark{2}Information and Computing Technologies (ICT) Division, College of Science and Engineering (CSE), Hamad Bin Khalifa University, Doha, Qatar.
    \\\{kahmad,aalfuqaha\}@hbku.edu.qa} \\
     \IEEEauthorblockA{\IEEEauthorrefmark{3}SimulaMet, Norway.
    \\\{steven,michael,paalh\}@simula.no}
}

\maketitle

\begin{abstract}
\label{abstract}
The increasing popularity of social networks and users' tendency towards sharing their feelings, expressions and opinions in text, visual, and audio content, have opened new opportunities and challenges in sentiment analysis. While sentiment analysis of text streams has been widely explored in literature, sentiment analysis from images and videos is relatively new. This article focuses on visual sentiment analysis in a societal important domain, namely disaster analysis in social media.
To this aim, we propose a deep visual sentiment analyzer for disaster-related images, covering different aspects of visual sentiment analysis starting from data collection, annotation, model selection, implementation, and evaluations. For data annotation, and analyzing people's sentiments towards natural disasters and associated images in social media, a crowd-sourcing study has been conducted with a large number of participants worldwide. The crowd-sourcing study resulted in a large-scale benchmark dataset with four different sets of annotations, each aiming a separate task. The presented analysis and the associated dataset will provide a baseline/benchmark for future research in the domain.
We believe the proposed system can contribute toward more livable communities by helping different stakeholders, such as news broadcasters, humanitarian organizations, as well as general public.

\end{abstract}

\section{Introduction}
Sentiment analysis aims to analyze and extract opinions, views, and perceptions about an entity (e.g., product, service, or an action). It has been widely adopted by businesses helping them to understand consumers' perceptions about their products and services. The recent development and popularity of social media helped researchers to extend the scope of sentiment analysis to other interesting applications. A recent example is reported by Ozturk et al.~\cite{ozturk2018sentiment} where computational sentiment analysis is applied to the leading media sources as well as social media to extract sentiments on the Syrian's refugee crisis. Another example is reported by Kuvsen et al.~\cite{kuvsen2018politics} where the neutrality of tweets and other reports from the winner of the Austrian presidential election were analyzed and compared to the opponents' content on social media.

The concept of sentiment analysis has been widely utilized in Natural Language Processing (NLP), where several techniques have been employed to extract sentiments from text streams in terms of positive, negative, and neutral perception/opinion. With the recent advancement in NLP, an in-depth analysis of text streams from different sources is possible in different application domains, such as education, entertainment, hosteling, and other businesses~\cite{sadr2019robust}. More recently, several efforts have been made to analyze visual contents to derive sentiments. The vast majority of literature on visual sentiment analysis focuses on facial close-up images where facial expressions are used as visual cues to derive sentiments and predict emotions~\cite{barrett2019emotional}. Attempts have been made also to extend the visual approach to more complex images, including, for example, multiple objects and background details. The recent developments in machine learning and in particular deep learning have contributed to significantly boost the results also in this research area~\cite{poria2018multimodal}. However, extracting sentiments from visual contents is not straightforward and several factors need to be considered.

In this article, we analyze the problem of visual sentiment analysis from different perspectives with a particular focus on the challenges, opportunities, and potential applications of visual sentiment analysis of challenging disaster-related images from social media. Disaster analysis in social media content received great attention of the community in recent years~\cite{said2019natural,imran2020using,ahmad2018social}. We believe visual sentiment analysis of disaster-related images is an exciting research domain that will benefit users and the community in a diversified set of applications. To this aim, we propose a deep sentiment analyzer, and discuss the processing pipeline of visual sentiment analysis starting from data collection and annotation via a crowd-sourcing study, and conclude with the development and training of deep learning models. The work is motivated by our initial efforts in the domain~\cite{hassan2019sentiment}, where an initial crowd-sourcing study was conducted with a few volunteers to test the viability of the approach. 

To the best of our knowledge, this is the first attempt to develop a large-scale benchmark for sentiment analysis of disaster-related visual content. Disaster-related images are complex and generally involve several objects as well as significant details in their backgrounds. We believe, such a challenging use-case is quintessential being an opportunity to discuss the processing pipeline of visual sentiment analysis and provide a baseline for future research in the domain. Moreover, visual sentiment analysis of disasters, has several applications and can contribute toward more livable communities. It can also help news agencies to cover such adverse events from different angles and perspectives. Similarly, humanitarian organizations can benefit from such a framework to spread the information on a wider scale, focusing on the visual content that best demonstrates the evidence of a certain event. In order to facilitate the future work in the domain, a large-scale dataset is collected, annotated, and made publicly available\footnote{https://datasets.simula.no/image-sentiment/}. For the annotation of the dataset, a crowd-sourcing activity with a large number of participants has been conducted.

The main contributions of the work can be summarized as follows:

\begin{itemize}
    \item We extend the concept of visual sentiment analysis to a more challenging and crucial task of disaster analysis, generally involving multiple objects and other relevant information in the background of images, and propose a deep architectures-based visual sentiment analyzer for an automatic sentiment analysis of natural disaster-related images from social media.
   \item Assuming that the available deep architectures respond differently to an image by extracting diverse but complementary image features, we evaluate the performance of several deep architectures pre-trained on ImageNet and Places dataset both individually and in combination.  
    \item We conduct a crowd-sourcing study to analyze people's sentiments towards disasters and disaster-related content, and annotate training data. In the study, a total of 2,338 users participated to analyze and annotate 4,003 disaster-related images\footnote{All images are Creative Commons licensed.}.
    \item We provide a benchmark visual sentiment analysis dataset with four different sets of annotations, each aimed at solving a separate task, which is expected to be proved as a useful resource for future work in the domain. To the best of our knowledge, this is the first attempt on the subject.
\end{itemize}

The rest of the paper is organized as following: Section~\ref{visual_sentiment:background} motivates the work by differentiating it from the related concepts, such as emotion and facial recognition as well as textual sentiment analysis, and emphasizing on the opportunities, challenges and potential applications. Section~\ref{sec:methodology} describes the proposed pipeline for visual sentiment analysis of natural disaster-related images. Section~\ref{sec:experiments} provides the statistics of the crowd-sourcing study along with the experimental results of the proposed deep sentiment analyzer. Section~\ref{conclusion} concludes this study and provides directions for future research.

\section{Motivation, Concepts, Challenges and Applications}\label{visual_sentiment:background}

As implied by the popular proverb \textit{"a picture is worth a thousand words,"} visual contents are an effective mean to convey not only facts but also cues about sentiments and emotions. Such cues representing the emotions and sentiments of the photographers may trigger similar feelings from the observer and could be of help in understanding visual contents beyond semantic concepts in different application domains, such as education, entertainment, advertisement, and journalism. To this aim, masters of photography have always utilized smart choices, especially in terms of scenes, perspective, angle of shooting, and color filtering, to let the underlying information smoothly flow to the general public. Similarly, every user aiming to increase in popularity on the Internet will utilize the same tricks~\cite{chua2016follow}. However, it is not fully clear how such emotional cues can be evoked by visual contents and more importantly how the sentiments derived from a scene by an automatic algorithm can be expressed. This opens an interesting line of research to interpret emotions and sentiments perceived by users viewing visual contents.   

In the literature, emotions, opinion mining, feelings, and sentiment analysis have been used interchangeably~\cite{munezero2014they,barrett2019emotional}. In practice, there is a significant difference among those terms. Sentiments are influenced by emotions, and they allow individuals to show their emotions through expressions. In short, sentiments can be defined as a combination of emotions and cognition. Therefore, sentiments reveal underlying emotions through ways that require cognition (e.g., speech, actions, or written content).  

\begin{figure}
  \includegraphics[height=0.30\linewidth,width=0.30\linewidth]{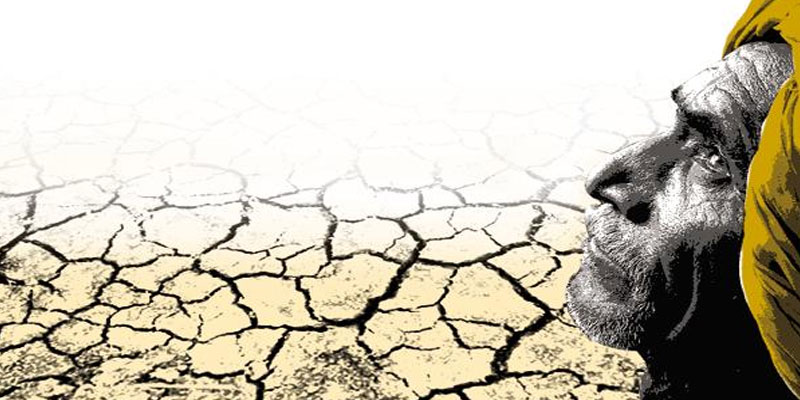}\hfill
  \includegraphics[height=0.30\linewidth,width=0.30\linewidth]{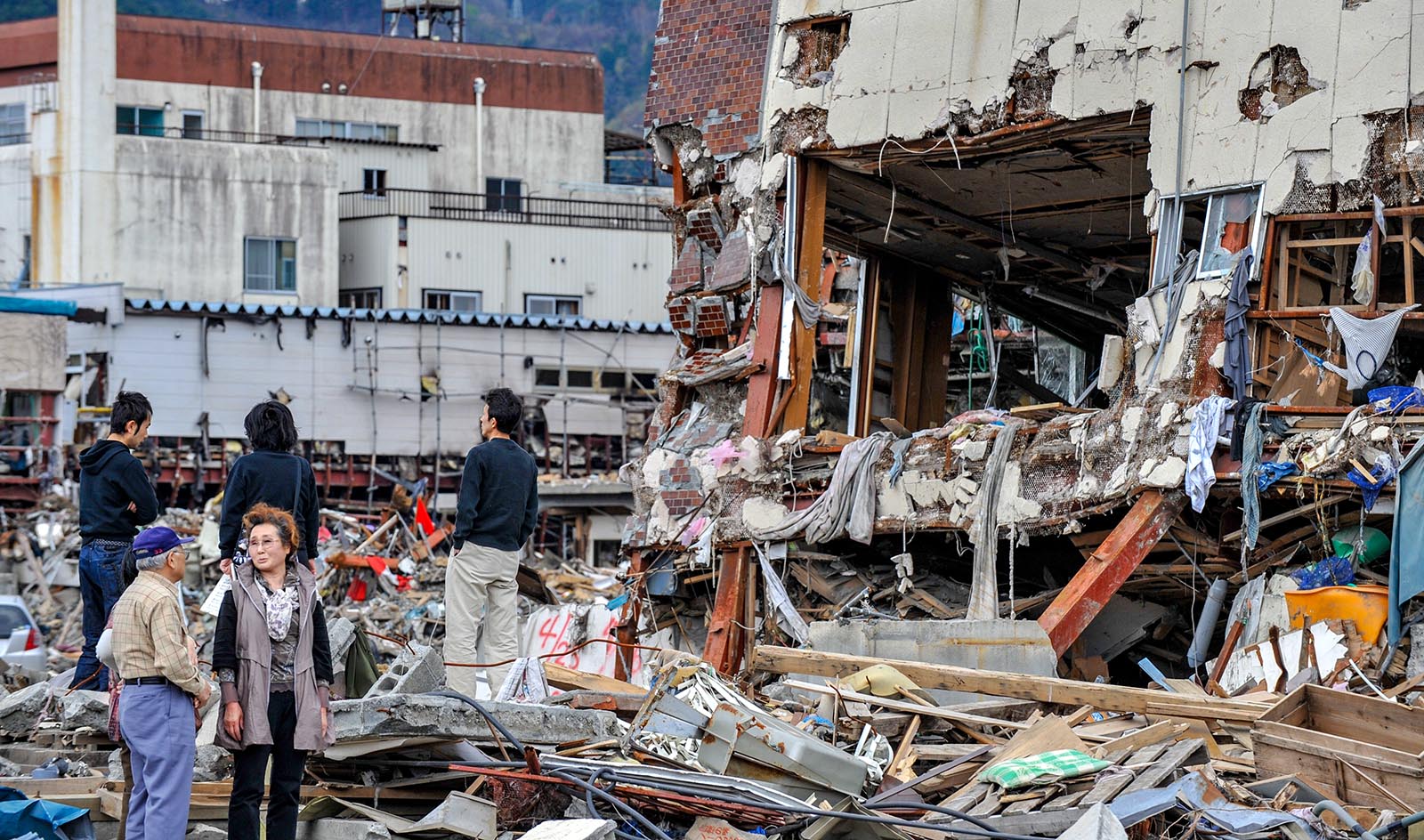}\hfill
  \includegraphics[height=0.30\linewidth,width=0.30\linewidth]{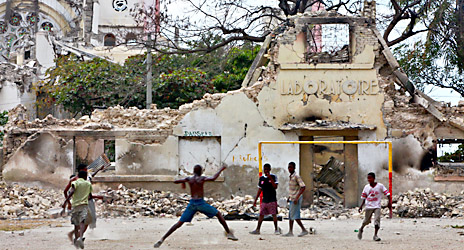}\hfill
  \caption{Sample images of natural disaster for sentiment analysis showing the diversity in the content and information to be extracted.}
  \label{fig:sample_images}
\end{figure}

The categorical representation of those concepts (i.e., emotions, sentiments) can be different, although the visual cues representing them are closely related. For instance, emotion recognition, opinion mining and sentiment analysis are generally expressed by three main classes: \textit{happy}, \textit{sad}, and \textit{neutral} or, similarly, \textit{positive}, \textit{negative}, and \textit{neutral}~\cite{kim2018building}. However, similar types of visual features are used to infer those states~\cite{soleymani2017survey}. For instance, facial expressions have been widely explored for both emotion recognition and visual sentiment analysis in close-up images~\cite{barrett2019emotional,khan2020face}; though it would be simplistic to limit the capability of recognizing emotions and sentiments to face-close up images. There are several application domains where more complex images need to be analyzed. This is exactly the case of the aforementioned scenario of disaster-related images, in which the background information is often crucial to evoke someone's emotions and sentiments. Figure~\ref{fig:sample_images} provides samples of disaster-related images, highlighting the diversity in terms of content that needs to be examined. In addition, it is also important to mention that emotions and feelings can be different from subject to subject, and based on experience. 

In contrast to textual sentiment analysis, visual sentiment analysis is a nascent area of research, and several aspects still need to be investigated. The following are some of the key open research challenges, in visual sentiment analysis in general and disaster-related content in particular, that need to be addressed:

\begin{itemize}
     
     \item \textbf{Defining/identifying sentiments} - The biggest challenge in this domain is defining sentiments and identifying the one that better suits given visual content. Sentiments are very subjective and vary from person to person. 
     Moreover, the intensity of the sentiments conveyed by an image is another item to be tackled. 
    
    \item \textbf{Semantic gap} - one of the open questions that researchers have thoroughly investigated in the past decades is the semantic gap between the visual features and the cognition~\cite{soleymani2017survey}. The selection of visual features is very crucial in multimedia analysis in general and in sentiment analysis in particular. We believe object and scene-level features could help in extracting such visual cues.
    
    
   \item \textbf{Data collection and annotation} - image sources, sentiments labels, and feature selection are application-dependent. For example, an entertainment or education context is completely different from the humanitarian one. Such diversity makes it difficult to collect benchmark datasets from which knowledge can be transferred, thus requiring ad-hoc data crawling and annotation. 
\end{itemize}


   
    

\section{Related work}\label{sec:related_work}
Natural language processing has made great strides in accurately determining the sentiment of a given spoken text, with reference to users' reviews on movies and products~\cite{10.1145/3041021.3054223, Araque2019, zhang2018deep}. When looking at the inference of sentiments from visual data, the literature is rather limited~\cite{Ortis2020}. However, being a new and challenging task, the lack of openly available datasets makes it difficult to create a common benchmark on which a solid state-of-the-art can be built. Machajdik et al.~\cite{machajdik2010affective} did a study on using extracted features based on psychology and art theory to classify the emotional response of a given image. The features were grouped by color, texture, composition, and content, and then classified by a naive Bayes-based classifier. Although the work achieved good results for the time, the extracted features have a hard time capturing the complex relationship between human emotion and the content of an image. Thus, more recent works have relied on reaching some middle-ground by extracting adjective-noun pairs (ANPs), like \textit{funny dog} or \textit{sad monkey}, which then may be used to infer the sentiment of the image. Borth et al.~\cite{10.1145/2502081.2502282} released a dataset consisting of over 3,000 ANPs, aimed to help researchers contributing to the field. Their work also includes a set of baseline models and is commonly used to benchmark methods based on ANPs~\cite{chen2014deepsentibank}. Another widely used approach that bypasses the need for large sentiment datasets consists of using deep neural networks and transfer learning from models trained on large-scale classification datasets like ImageNet~\cite{deng2009imagenet}. Al-Halah et al.~\cite{AlHalah2019SmileBH} developed a method for predicting emoticons (emojis) based on a given image. The emojis act as a proxy for the emotional response of an image. They collected a dataset containing over 4 million images and emoticon pairs from Twitter, which was used to train a novel CNN architecture named SimleyNet~\cite{AlHalah2019SmileBH}. Some works also employed the text associated with images for visual sentiment analysis. For instance, in \cite{10.1145/3388861}, an attention-based network, namely Attention-based Modality-Gated Networks (AMGN), has been proposed to exploit the correlation between visual and textual information for sentiment analysis. 


\section{Proposed Visual Sentiment Analysis Processing Pipeline}
\label{sec:methodology}

\begin{figure*}[!ht]
    \centering
	\includegraphics[width=0.99\textwidth]{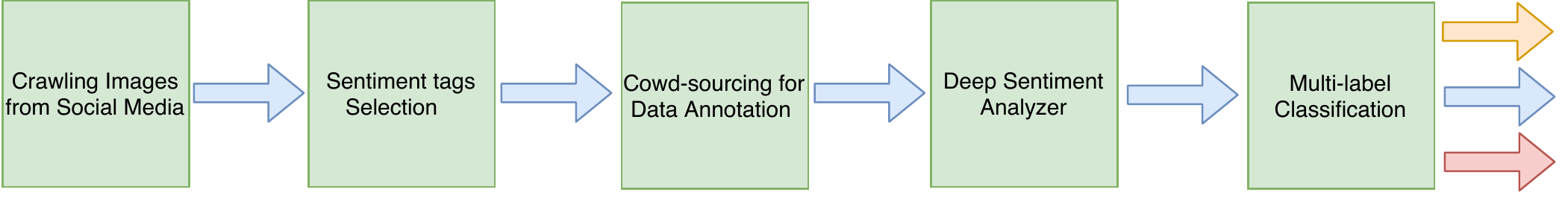}
	\caption{Block diagram of the proposed visual sentiment analysis processing pipeline.}
	\label{fig:methodology}
\end{figure*}

Figure~\ref{fig:methodology} provides the block diagram of the proposed architecture for visual sentiment analysis. The pipeline is composed of five phases. The process starts with crawling social media platforms for disaster-related images, followed by sentiment tags/categories selection to be associated with the disaster-related images in the crowd-sourcing study. Before conducting the crowd-sourcing study, we manually analyzed the images, and removed irrelevant images. In the crowd-sourcing study, a subset of the downloaded images, after removing the irrelevant images, are annotated with human participants. A CNN and a transfer learning-based method are then used for multi-label classification and to automatically assign sentiments/tags to the images. In the next subsections, we provide a detailed description of each component.

\subsection{Data collection and sentiment category selection}
At the beginning of the processing pipeline, social media platforms, such as Twitter and Flickr, and Google API are crawled to collect images to analyze. All the downloaded images have been selected paying attention to the licensing policies, in terms of free usage and sharing. The images have been selected according to a set of keywords, such as \textit{floods}, \textit{hurricanes}, \textit{wildfires}, \textit{droughts}, \textit{landslides} and \textit{earthquakes}, and enriched, with additional relevant information, as, for example, location (\textit{cyclones in Fiji} or \textit{floods in Pakistan}), and accessing the list of recent natural disasters made available from EM-DAT\footnote{https://www.emdat.be/}. EM-DAT is a platform maintained by the United Nations providing statistics on worldwide disasters. 

The selection of labels for the crowd-sourcing study was one of the challenging and perhaps most crucial phases of the work as discussed earlier. In literature, sentiments are generally represented as \textit{Positive}, \textit{Negative} and \textit{Neutral}~\cite{soleymani2017survey}. However, considering the nature and potential applications of the proposed deep sentiment analysis processing pipeline, we aim to target sentiments that are more specific to disaster-related contents. For instance, terms like \textit{sadness}, \textit{fear}, and \textit{destruction} are more commonly used with disaster-related contents. In order to choose more relevant and representative labels for our deep sentiment analyzer, we choose four different sets of tags including the most commonly used one, and two sets obtained from a recent work in Psychology~\cite{cowen2017self}, reporting $27$ different types of emotions. In total, we annotated every image with four different sets of labels. The first set is composed of three tags namely \textit{positive}, \textit{negative}, and \textit{neutral}. The second set also contains three tags, namely \textit{relax/calm}, \textit{normal}, and \textit{stimulated/excited}. The third set is composed of seven tags, namely \textit{joy}, \textit{sadness}, \textit{fear}, \textit{disgust}, \textit{anger}, \textit{surprise}, and \textit{neutral}. The last set of tags is composed of ten tags, namely \textit{anger}, \textit{anxiety}, \textit{craving}, \textit{empathetic pain}, \textit{fear}, \textit{horror}, \textit{joy}, \textit{relief}, \textit{sadness}, and \textit{surprise}. Table \ref{sets_tags} lists the tags used in each question of the crowd-sourcing. The basic motivation for the four different sets of labels is to cover different aspects of the task, and analyze how the complexity of the task varies by going deeper in the sentiments hierarchy.

 \begin{table}[!h]
 \centering
\begin{tabular}{|p{.9 cm}|p{7 cm}|}
\hline
\textbf{Sets} & \textbf{Tags} \\ \hline
      Set 1        &    \textit{Positive}, \textit{Negative}, \textit{Neutral}  \\ \hline
      Set 2        &  \textit{Relax}, \textit{Stimulated}, \textit{Normal}   \\ \hline
       Set 3       &   \textit{Joy}, \textit{Sadness}, \textit{Fear}, \textit{Disgust}, \textit{Anger}, \textit{Surprise}, and \textit{Neutral}            \\ \hline
        Set 4       &   \textit{Anger}, \textit{Anxiety}, \textit{craving}, \textit{Empathetic pain}, \textit{Fear}, \textit{Horror}, \textit{Joy}, \textit{Relief}, \textit{Sadness}, and \textit{Surprise}            \\ \hline
\end{tabular}
\caption{List of tags used in the crowd-sourcing study in the four sets.}
\label{sets_tags}
\end{table}

\subsection{The crowd-sourcing study}\label{sec:crowd-sourcing}
The crowd-sourcing study aims to develop ground-truth for the proposed deep sentiment analyzer by collecting human perceptions and sentiments about disasters and associated visual contents. The crowd-sourcing study was conducted using Microworkers \footnote{https://www.microworkers.com}, where the selected images were presented to the participants to be annotated. In total, 4,003 images were analyzed during the study. In order to assure the quality of the annotations, at least five different participants are assigned to analyze an image. The final tag/tags are chosen based on the majority votes from the five participants assigned to it. In total, 10,010 different responses were obtained during the study from 2,338 different participants. The participants included individuals from different age groups and 98 different countries. We also noted the time spent by a participant on an image, which helped in filtering out the careless or inappropriate responses from the participants. The average response time recorded per image during the study is 139 seconds.
Before the final study was conducted, two trial studies were performed to fine tune the test, correct errors, and improve clarity and readability. The HTML version of the crowdsourcing study template has been made available as a part of the dataset.

\begin{figure}[!ht]
    \centering
    \includegraphics[width=0.99\linewidth]{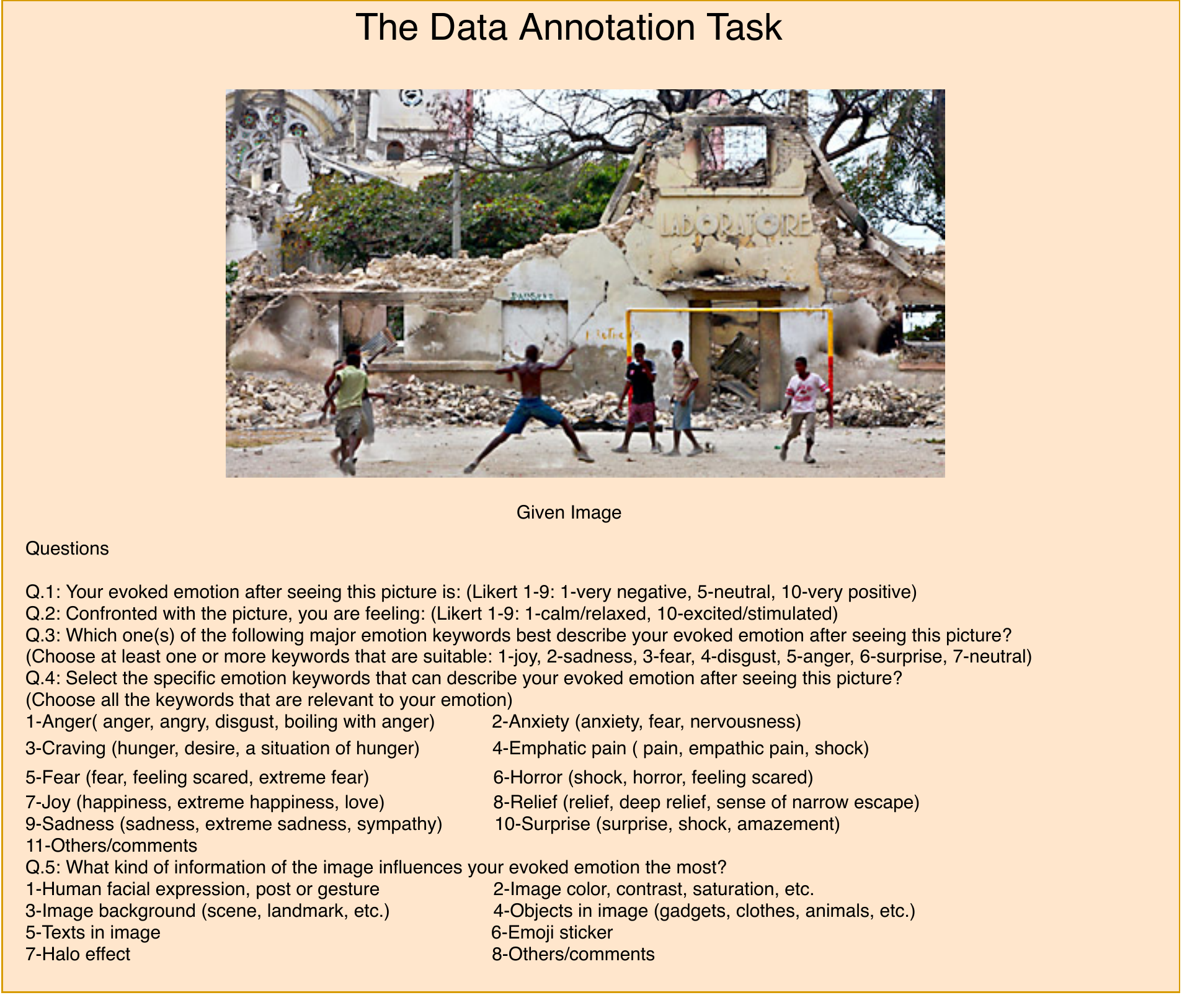}
    \caption{An illustration of the web application used for the crowd-sourcing study. A disaster-related image is provided to the users who are asked to provide options/tags. In case, additional tags/comments can also be reported.}
	\label{crowdsourcing_study}
\end{figure}

Figure~\ref{crowdsourcing_study} provides a block diagram of the layout of the crowd-sourcing study platform. The participants were provided with a disaster-related image followed by five different questions. In the first four questions, the participants are asked to annotate the image with different sets of labels (\ref{sets_tags}) each aiming to prepare training data for a separate task. In the first question, we asked the participants to rate their evoked emotions from 1 to 10 (1-very negative, 5-neutral, and 10-very positive), after seeing the image. This question aims to analyze the degree of sentiments conveyed by an image. The second question is similar to the first one except the labels focus on feelings in terms of \textit{calm/relaxed}, \textit{normal}, and \textit{excited}. In the third and fourth questions, the participants are asked to assign one or more label from a list of seven and ten tags, respectively. In these two questions, the participants were also encouraged to provide their own tags if they felt the provided lists are not representative enough. 
The fifth question aims to highlight the image features, at scene or object level, which influence human emotions. 

The resulting dataset is named image-sentiment dataset and can be downloaded via \url{https://datasets.simula.no/image-sentiment/}. Details about the dataset can be found in Table \ref{dataset_task1}, Table \ref{dataset_task2}, and Table \ref{dataset_task3}.

\subsection{Deep Visual Sentiment Analyzer}
Our proposed multi-label deep visual sentiment analyzer is mainly based on a convolutional neural network (CNN) and transfer learning. Based on the participants' responses in the fifth question, where they were asked to highlight the image features/information that influence their emotions and sentiments, we believe both object and scene-level features could be useful for the classification task. Thus, we opted for both object and scene-level features extracted through existing deep models pre-trained on the ImageNet~\cite{deng2009imagenet} and Places~\cite{zhou2014learningB} datasets, respectively. The model pre-trained on ImageNet extracts object-level information, while the one pre-trained on the Places dataset covers the background details \cite{ahmad2019deep}. In this work, we employed several state-of-the-art deep models, namely AlexNet~\cite{krizhevsky2012imagenet}, VGGNet~\cite{simonyan2014very}, ResNet~\cite{he2016deep}, Inception v3~\cite{szegedy2016rethinking}, DenseNet~\cite{iandola2014densenet}, and EfficientNet~\cite{tan2019efficientnet}. These models are fine-tuned on the newly collected dataset for visual sentiment analysis of disaster-related images. The object and scene-level features are also combined using early fusion by including a concatenation layer in our framework, where features from models pre-trained on the ImageNet and Places datasets are combined before the classification layer. In the current implementation, we rely on a simple fusion technique aiming to identify and analyze the potential improvement by combining both object and background details. 
In addition, in order to deal with class imbalance as will be detailed in Section~\ref{sec:dataset}, we also used an oversampling techniques to adjust the class distribution of the dataset. For the single-label classification (i.e., first task), we used an open-source library, namely \textit{imblearn}~\cite{JMLR:v18:16-365}, while for the multi-label problem (i.e., second and third task) we developed our own function. In fact, in the multi-label tasks the classes are not independent, thus the naïve approach (i.e., \textit{imblearn}) could not be applied. In order to deal with it, we divided the classes into two groups based on positive and negative correlation with the majority class occurrence. Then, each group is sorted in descending order based on the number of samples in each class. We then iterate over the each group, and oversample the minority classes. 

Furthermore, for the multi-label analysis, we made several changes in the framework to adapt the pre-trained models for the task at hand. As a first step, a vector of the ground truth having all the possible labels has been created with the corresponding changes in the models. For instance, the top layer of the model has been modified to support multi-label classification by replacing the soft-max function with a sigmoid function. The sigmoid function turns out to be helpful as it presents the results for each label in probabilistic terms, while the soft-max function holds the probability law and squashes all the values of a vector into a \textit{[0,1]} range. Similar changes (i.e., replacing softmax with sigmoid function) are made in the formulation of the cross-entropy to properly fine-tune the pre-trained models. During the experiments, we used 70\% data for training, 10\% for validation, and 20\% for test purposes. The experiments are carried out on Intel(R) machine Core(TM) i7-8700 with GPU GeForce RTX 2070 (8GB) and 62GB of RAM. 

\section{Experiments and Evaluations}
\label{sec:experiments}

In this section, we provide a detailed analysis of the outcomes of the crowed-sourcing study and achieved experimental results.

\subsection{Statistics of the Crowd-sourcing study and Dataset}
\label{sec:crowd-sourcing_statistics}
\begin{figure*}[!h]
    \centering
    \begin{subfigure}{0.45\linewidth}
        \includegraphics[width=\linewidth]{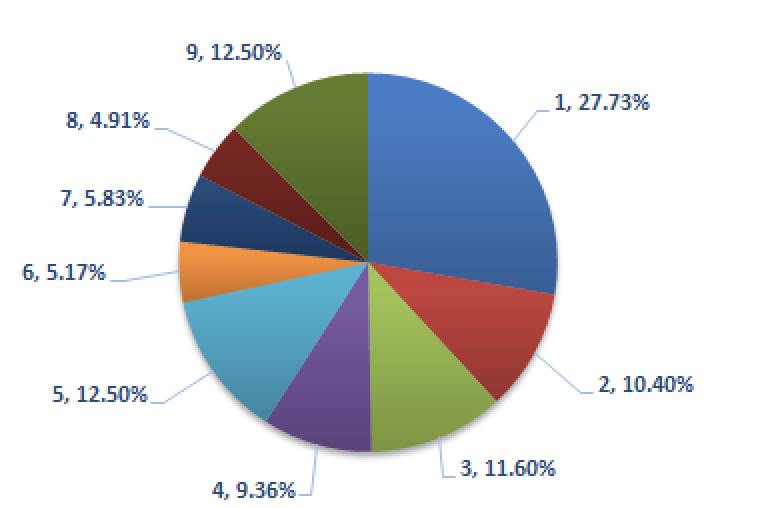}
        \caption{Statistics of the responses for the first question. Tags 1 to 4 represent negative sentiments while tag 5 represents neutral, and tags 6 to 9 show positive sentiments.}\label{fig:a}
    \end{subfigure}
    \begin{subfigure}{0.45\linewidth}
        \includegraphics[width=\linewidth]{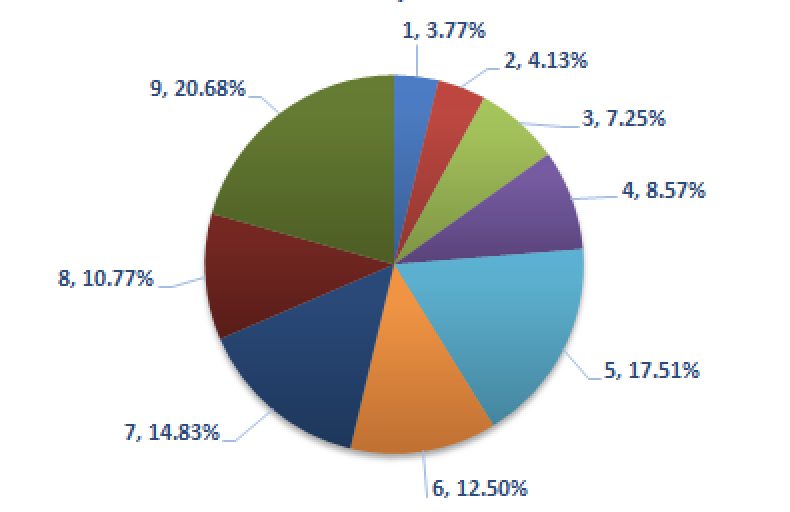}
        \caption{Statistics of the responses for the second question. Tags 1 to 4 represent calm/relaxed emotion, tag 5 shows normal condition while tags 6 to 9 depict excited/stimulated status.}\label{fig:b}
    \end{subfigure}
    
    \begin{subfigure}{0.45\linewidth}
        \includegraphics[width=\linewidth]{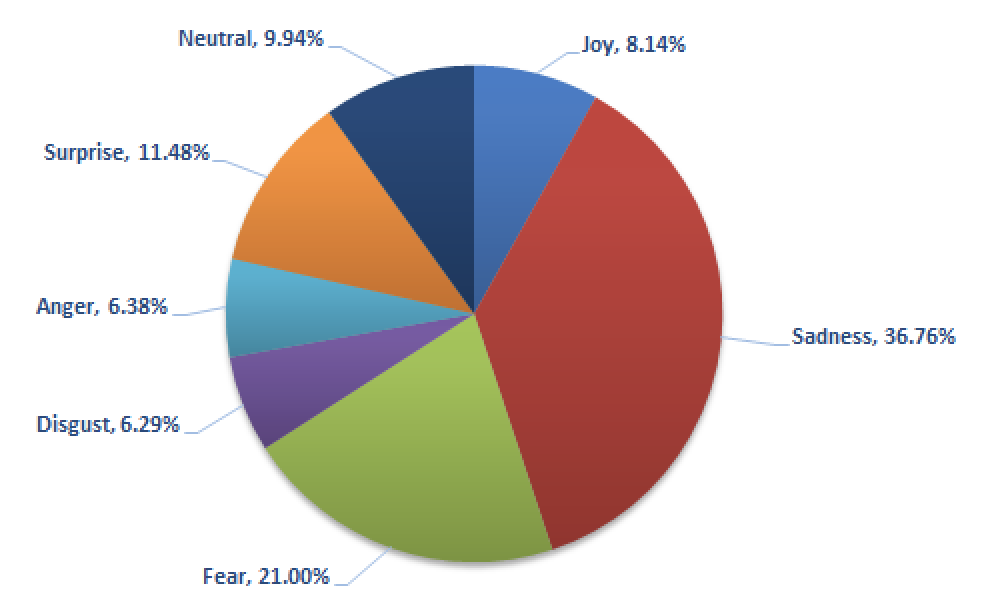}
        \caption{Statistics of the responses for the third question.}\label{fig:c}
    \end{subfigure}
    \begin{subfigure}{0.45\linewidth}
        \includegraphics[width=\linewidth]{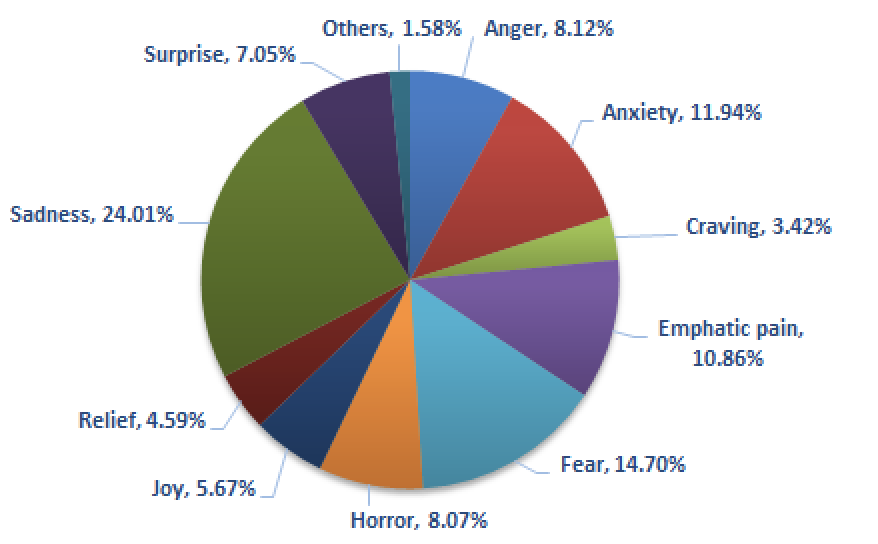}
        \caption{Statistics of the responses for the fourth question.}\label{fig:d}
    \end{subfigure}
    \caption{Statistics of the first four questions of the crowd-sourcing study.}
    \label{fig:cs_statistics}
\end{figure*}

    Figure~\ref{fig:cs_statistics} provides the statistics of the first four questions of the crowd-sourcing study. Figure~\ref{fig:a} (where tags 1 to 4 correspond to \textit{negative} sentiments, 5 to \textit{neutral}, and tags 6 to 9 represent \textit{positive} sentiments) shows that majority of the images analyzed in the crowd-sourcing study evoked \textit{negative} sentiments. 
    Looking at the remaining responses, we noticed that images labelled as \textit{positive} are mostly captured during the rescue, and rehabilitation process. Figure~\ref{fig:b} provides the statistics of the second question, which is based on a different set of labels: \textit{calm}/\textit{relaxed}, \textit{normal}, and \textit{stimulated}/\textit{excited}. The emotions are here quite evenly distributed across the entire spectrum ranging from negative to positive. 
    Figure~\ref{fig:c} provides the statistics of the third question of the study in terms of how frequently different tags are assigned with the images by the participants. As expected, \textit{sadness} and \textit{fear} are the most frequently used tags. The statistics of Question 4, which further extends the tags' list by going deeper in the hierarchy, are shown in Figure~\ref{fig:d}. Similar to Question 3, higher percentages have been observed for \textit{sadness} and \textit{fear}.

\begin{figure}[!h]
\begin{subfigure}{\linewidth}
\centering
\includegraphics[width=.99\linewidth]{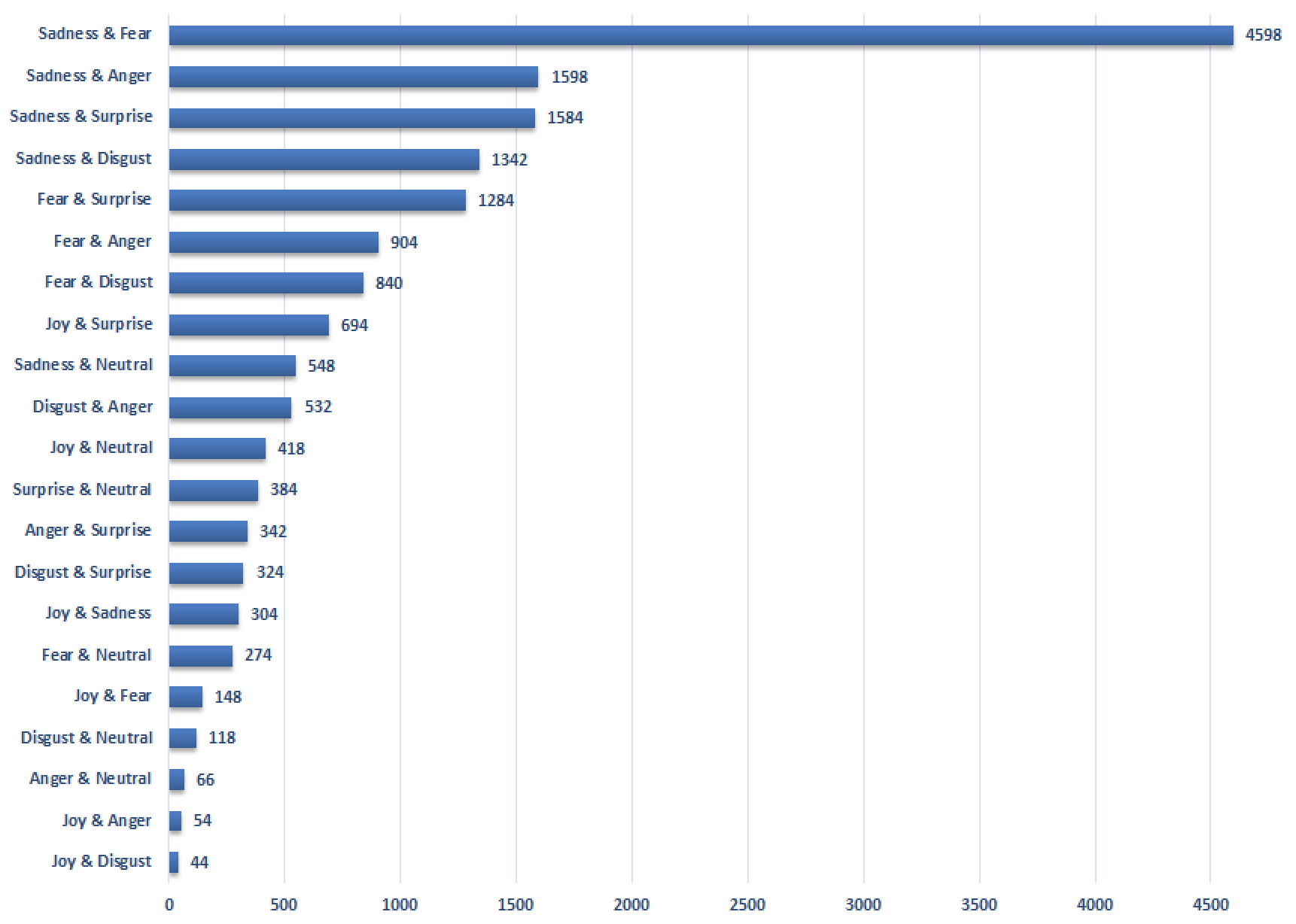}
\caption{Tags jointly used in pairs.}
\label{cs_statistics_pairs}
\vspace{3mm}
\end{subfigure}
\begin{subfigure}{\linewidth}
\centering
\includegraphics[width=.99\linewidth]{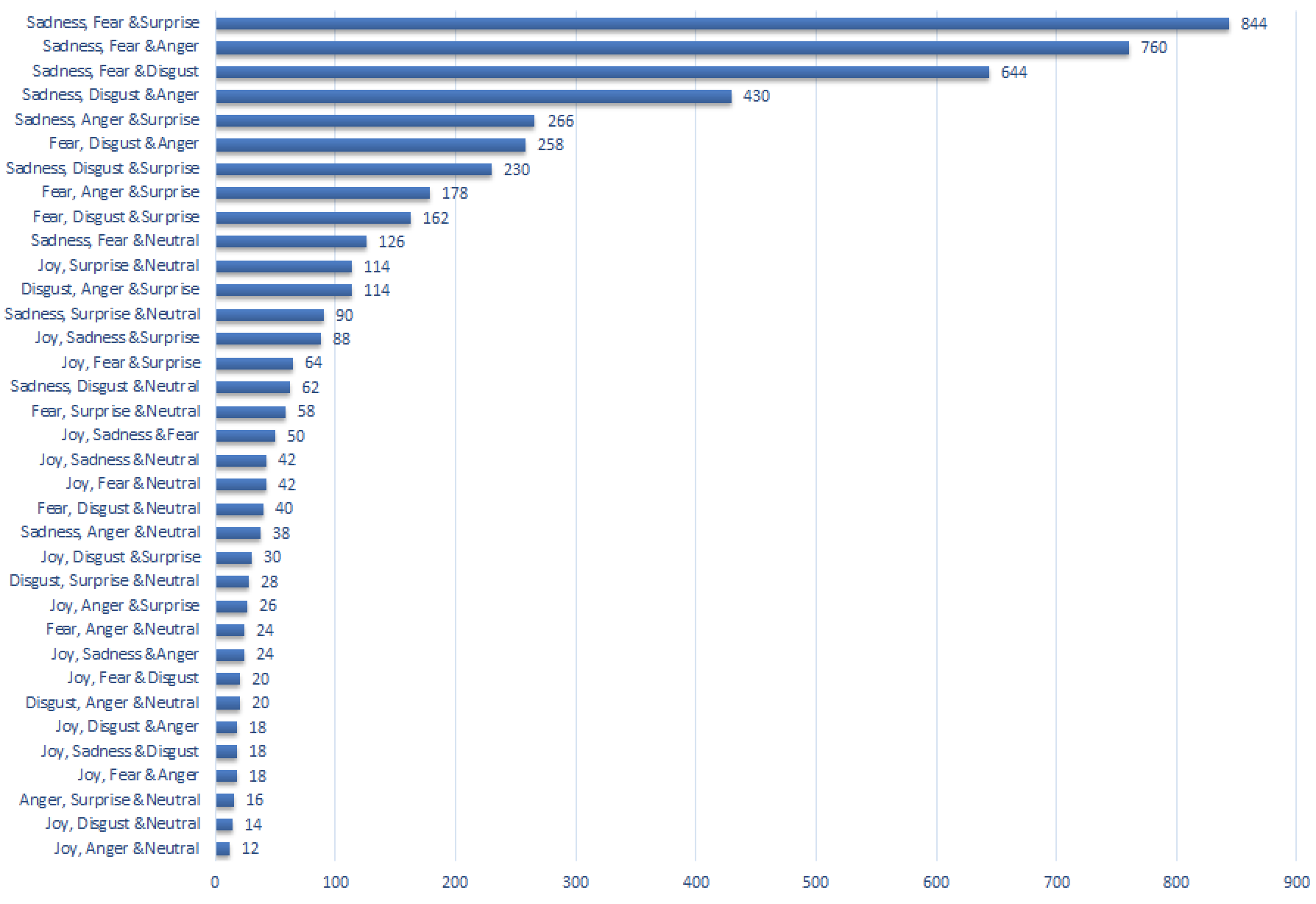}
\caption{Tags jointly used in group of three.}
\label{cs_statistics_G3}
\end{subfigure}
\caption{Statistics of the crowd-sourcing study in terms of how different tags are jointly associated images.}
\end{figure}

Another important aspect of the crowd-sourcing study is he analysis related to how frequently different tags are used jointly. In Figure~\ref{cs_statistics_pairs} and Figure~\ref{cs_statistics_G3}, we show this association in pairs, and groups of three, respectively. As can be seen, \textit{sadness} is most frequently used with \textit{fear}, \textit{anger}, and \textit{disgust}. Similarly, \textit{fear} is also frequently used with \textit{disgust}, \textit{anger}, \textit{surprise}. As for the positive tags, \textit{joy}, \textit{surprise} and \textit{neutral} are jointly used. A similar trend has been observed in the group of three tags. 

\begin{figure}[!h]
\centering
\includegraphics[width=0.9\linewidth]{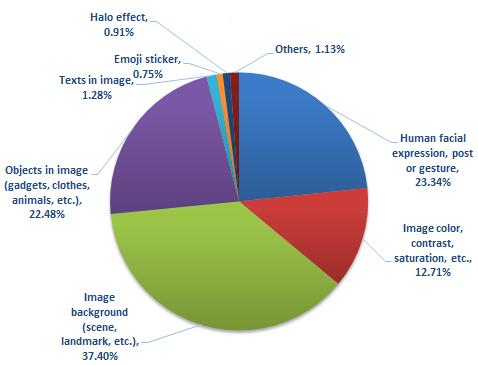}
\caption{Statistics of the fifth question of the crowd-sourcing study in terms of what kind of information in the images influence users' emotion most.}
	\label{cs_statistics_q5}
\end{figure}

Figure~\ref{cs_statistics_q5} provides the statistics of the final question of the study, where we asked the participants to highlight the image features/information that influence their emotions and tag selection for a given image. This question is expected to provide useful information from a methodological point of view. As can be seen, the image background (i.e., scene, landmarks, etc.,) has been proved the most influential piece of information for evoking people's emotions (37.40\% of the responses). Human expressions, gestures, and poses also seem very crucial (23\%). Other factors, such as object-level information in images, and image color and contrast, contributed with 22.48\% and 12.71\%, respectively.

\subsection{Datasets}
\label{sec:dataset}
In this section, we provide the details of the datasets we have collected and adopted for the crowdsourcing study. Table~\ref{dataset_task1}, Table~\ref{dataset_task2}, and Table~\ref{dataset_task3} report the statistics of the dataset, in terms of total number of samples per class, used for each of the three tasks. For the first task images are arranged in three different classes, namely \textit{positive}, \textit{negative}, and \textit{neutral} with a bias towards the negative samples, due to the topic taken into consideration. For this task the dataset is single-label. In task 2 and task 3 we use instead a multi-label annotation, and the dataset contains images from seven and ten different classes, respectively. As can be seen in Table~\ref{dataset_task2}, the majority of the classes have a higher number of images, and some of the classes have a similar range. For instance, \textit{anger}, \textit{joy}, and \textit{disgust} have samples in the same range while \textit{neutral} and \textit{surprise}, and \textit{sadness} and \textit{fear} have almost the same amount of samples. One of the reasons for the pattern is the joint association of the tags with images by the participants of the crowd-sourcing study. Similar pattern can be observed in Table~\ref{dataset_task3} for task 3, where sentiment classes, such as \textit{craving}, \textit{joy}, and \textit{relief} have number of samples in the same range. Similarly, \textit{anger}, \textit{horror}, and \textit{surprise} have the same range of number of samples. On the other hand, the number of samples in \textit{fear}, \textit{sadness}, and \textit{anxiety} are in the same range.

\begin{table}[!h]
\caption{Statistics of the Datasets used in all tasks}
\begin{subtable}[c]{0.5\textwidth}
\centering
\begin{tabular}{|c|c|}
\hline
\textbf{Tags} & \textbf{\# Samples}\\ \hline
Positive &  803\\ \hline
Negative & 2297\\ \hline
Neutral & 579\\ \hline
\end{tabular}
\caption{Statistics of the dataset for the task 1.}
\label{dataset_task1}
\vspace{4mm}
\end{subtable}
\begin{subtable}[c]{0.5\textwidth}
\centering
\begin{tabular}{|c|c|c|c|}
\hline
\textbf{Tags} & \textbf{\# Samples} & \textbf{Tags} & \textbf{\# Samples} \\ \hline
Joy& 1207 & Sadness & 3336\\ \hline
Fear& 2797& Disgust& 1428\\ \hline
Anger& 1419& Surprise& 2233\\ \hline
Neutral& 1892& - & - \\ \hline
\end{tabular}
\caption{Statistics of the dataset for the task 2.}
\label{dataset_task2}
\vspace{5mm}
\end{subtable}
\begin{subtable}[c]{0.5\textwidth}
\centering
\begin{tabular}{|c|c|c|c|}
\hline
\textbf{Tags} & \textbf{\# Samples} & \textbf{Tags} & \textbf{\# Samples} \\ \hline
Anger& 2108  &Anxiety & 2716\\ \hline
Craving& 1100 & Pain& 2544\\ \hline
Fear&2803 & Horror&2042 \\ \hline
Joy&1181 & Relief &1356 \\ \hline
Sadness& 3300& Surprise&1975 \\ \hline
\end{tabular}
\caption{Statistics of the dataset for the task 3.}
\label{dataset_task3}
\end{subtable}

\end{table}

\subsection{Experimental Results}

Table~\ref{results_task1}, Table~\ref{results_task2}, and Table~\ref{results_task3} provide experimental results of the proposed deep sentiment analyzer on the three tasks. Since one of the main motivations behind the experiments is to provide a baseline for future work in the domain, we evaluate the proposed single and multi-label frameworks with several existing deep models pre-trained on ImageNet and Places datasets.

For the first task, we evaluate the performance of the proposed single-label framework with several state-of-the-art models in differentiating in \textit{positive}, \textit{negative}, and \textit{neutral} sentiments. As shown in Table~\ref{results_task1}, we obtain encouraging results, in terms of accuracy, recall, precision, and F1-score. Surprisingly, better results have been observed for the smaller architecture (VGGNet) compared to the most recent models, such as EfficientNet and DenseNet. As far as the contribution of object and scene-level features is concerned, both types of features could turned out to be useful for the classification task. We also combined the object and scene-level feature following an early fusion approach; no significant improvement has been observed. 

Table~\ref{results_task2} provides the experimental results of the proposed multi-label framework, where the system needs to automatically associate to an image one or more labels from seven tags, namely sadness, fear, disgust, joy, anger, surprise, and neutral. Also in this case the results are encouraging especially considering the complexity of the tags in terms of inter and intra-class variation. In this case, the fusion of object and scene-level features outperforms the individual models in terms of accuracy, recall, and F1-score.

Table~\ref{results_task3} provides the results of the third task where we go deeper in the sentiments hierarchy with a total of ten tags. Also similar to previous tasks, the results are also encouraging on the more complex task where the sentiments' categories are increased further. 
\begin{table}[!h]
\caption{Evaluation of the proposed visual sentiment analyzer with different deep learning models pre-trained on the ImageNet and Places datasets.}
\begin{subtable}[c]{0.5\textwidth}
\scalebox{0.9}{
\begin{tabular}{|c|c|c|c|c|}
\hline
\textbf{Model} & \textbf{Accuracy} & \textbf{Precision } & \textbf{Recall } & \textbf{F-Score} \\ \hline

VGGNet (ImageNet) & 92.12 & 88.64 & 87.63 & 87.89 \\ \hline
VGGNet (Places)   & 92.88 & 89.92 & 88.43 & 89.07 \\ \hline
 Inception-v3 (ImageNet) & 82.59 & 76.38 & 68.81 & 71.60 \\ \hline
ResNet-50 (ImageNet) & 90.61 & 86.32 & 85.18 & 85.63 \\ \hline
ResNet-101 (ImageNet) & 90.90 & 86.79 & 85.84 & 86.01 \\ \hline
DenseNet (ImageNet) & 85.77 & 79.39 & 78.53 & 78.20 \\ \hline
EfficientNet (ImageNet) & 91.31 & 87.00 & 86.94 & 86.70  \\ \hline
VGGNet (places + ImageNet)  & 92.83 & 89.67 & 88.65 & 88.97 \\ \hline
\end{tabular}}
\caption{Evaluation of the proposed visual sentiment analyzer on the task 1 (i.e., single label classification of three classes, namely negative, neutral and positive).}
\label{results_task1}
\vspace{4mm}
\end{subtable}
\begin{subtable}[c]{0.5\textwidth}
\scalebox{0.9}{
\begin{tabular}{|c|c|c|c|c|}
\hline
\textbf{Model} & \textbf{Accuracy } & \textbf{Precision } & \textbf{Recall } & \textbf{F-Score } \\ \hline
VGGNet (ImageNet) & 82.61 & 84.12 & 80.28 & 81.66 \\ \hline
VGGNet (Places)   & 82.94 & 82.87 & 82.30 & 82.28 \\ \hline
 Inception-v3 (ImageNet) & 80.67 & 80.98 & 82.98 & 80.72 \\ \hline
ResNet-50 (ImageNet) & 82.48 & 84.33 & 79.41 & 81.38 \\ \hline
ResNet-101 (ImageNet) & 82.70 & 82.92 & 82.04 & 82.20 \\ \hline
DenseNet (ImageNet) & 81.99 & 83.43 & 81.30 & 81.51 \\ \hline
EfficientNet (ImageNet) & 82.08 & 82.80 & 81.31 & 81.51 \\ \hline
 VGGNet (places + ImageNet)    & 83.18 & 83.13 & 83.04 & 82.57 \\ \hline
\end{tabular}}
\caption{Evaluation of the proposed visual sentiment analyzer on the task 2 (i.e., multi-label classification of seven classes, namely sadness, fear, disgust, joy, anger, surprise, and neutral.}
\label{results_task2}
\vspace{5mm}
\end{subtable}
\begin{subtable}[c]{0.5\textwidth}
\scalebox{0.9}{
\begin{tabular}{|c|c|c|c|c|}
\hline
\textbf{Model} & \textbf{Accuracy } & \textbf{Precision } & \textbf{Recall } & \textbf{F-Score } \\ \hline
VGGNet (ImageNet) & 82.74 & 80.43 & 85.61 & 82.14 \\ \hline
VGGNet (Places)   & 81.55 & 79.26 & 85.08 & 81.16 \\ \hline
 Inception-v3 (ImageNet) & 81.53 & 78.21 & 89.30 & 82.27 \\ \hline
ResNet-50 (ImageNet) & 82.30 & 79.90 & 84.18 & 81.60 \\ \hline
ResNet-101 (ImageNet) & 82.56 & 80.25 & 84.51 & 81.80 \\ \hline
DenseNet (ImageNet) & 81.72 & 79.40 & 85.35 & 81.63 \\ \hline
EfficientNet (ImageNet) & 82.25 & 80.83 & 82.70 & 81.39 \\ \hline
 VGGNet (places + ImageNet)    & 82.08 & 79.36 & 87.25 & 81.99 \\ \hline
\end{tabular}}
\caption{Evaluation of the proposed visual sentiment analyzer on the task 3 (i.e., multi-label classification of seven classes, namely anger, anxiety, craving,  empathetic pain, fear, horror, joy, relief, sadness, and surprise.}
\label{results_task3}
\end{subtable}
\end{table}

For completeness, we also provide experimental results of the proposed methods in terms of accuracy, precision, recall, and F1-score per class. Table~\ref{results_per_class_task1} provides the experimental results of task 1. 
Looking at the performances of the model on the individual classes, we can notice that some have performed comparably better than others. For instance, on the \textit{positive} class, VGGNet pre-trained on Places dataset performed better compared to the all the other models.      

\begin{table}[]
\centering
\begin{tabular}{|c|c|c|c|c|}
\hline
\textbf{Model} & \textbf{Metric}  & \textbf{Negative} & \textbf{Neutral} & \textbf{Positive} \\ \hline
\multirow{4}{*}{VGGNet} & \textit{Accuracy}  & 88.61 &	95.36 & 91.66 \\ \cline{2-5} 
 & \textit{Precision} & 88.45 &	93.20 &	84.56 \\ \cline{2-5} 
 & \textit{Recall} & 74.59 &	93.29 &	91.83 \\ \cline{2-5} 
 & \textit{F1-Score} & 80.85 &	93.22 &	88.04 \\ \hline
\multirow{4}{*}{VGGNet (p)} & \textit{Accuracy}  & 90.07 &	94.88 & 93.21 \\ \cline{2-5} 
 & \textit{Precision} & 88.63 &	91.13 &	89.87 \\ \cline{2-5} 
 & \textit{Recall} & 79.52 &	94.21&	89.88 \\ \cline{2-5} 
 & \textit{F1-Score} & 83.79 &	92.64 &	89.85 \\ \hline
  \multirow{4}{*}{Inception V-3} &\textit{Accuracy}  & 76.48 &	86.51 &	82.28 \\ \cline{2-5} 
 & \textit{Precision} & 70.64 &	79.34 &	78.25 \\ \cline{2-5} 
 & \textit{Recall} & 45.76 &	82.51 &	66.86 \\ \cline{2-5} 
 & \textit{F1-Score} & 55.46 &	80.85 &	71.41 \\ \hline
 \multirow{4}{*}{ResNet-50} & \textit{Accuracy}  & 86.95 &	92.22 & 92.07 \\ \cline{2-5} 
 & \textit{Precision} & 83.40 &	87.15 &	88.14 \\ \cline{2-5} 
 & \textit{Recall} & 74.51 &	90.68 &	88.29 \\ \cline{2-5} 
 & \textit{F1-Score} & 78.65 &	88.86 &	88.170 \\ \hline
 \multirow{4}{*}{ResNet-101} & \textit{Accuracy}  & 87.16 &	92.31 & 92.29 \\ \cline{2-5} 
 & \textit{Precision} & 86.57 &	86.07 &	87.99 \\ \cline{2-5} 
 & \textit{Recall} & 71.38 &	92.80 &	89.15 \\ \cline{2-5} 
 & \textit{F1-Score} & 78.11 &	89.25 &	88.54 \\ \hline
 \multirow{4}{*}{DenseNet} & \textit{Accuracy}  & 80.59	& 87.84 &	87.72 \\ \cline{2-5} 
 & \textit{Precision} & 76.98 &	80.33	& 83.04 \\ \cline{2-5} 
 & \textit{Recall} & 60.16	& 87.01 &	79.54 \\ \cline{2-5} 
 & \textit{F1-Score} & 66.15 &	83.10 &	81.18 \\ \hline
 \multirow{4}{*}{EfficientNet} & \textit{Accuracy}  & 87.50	& 93.91 &	91.66 \\ \cline{2-5} 
 & \textit{Precision} & 86.41 &	93.91 &	84.87 \\ \cline{2-5} 
 & \textit{Recall} & 72.87 &	92.58 &	91.68 \\ \cline{2-5} 
 & \textit{F1-Score} & 78.96 &	91.24 &	88.07 \\ \hline
  \multirow{4}{*}{VGGNet (P+I)} & \textit{Accuracy}  & 89.94 &	94.90 &	92.99 \\ \cline{2-5} 
 & \textit{Precision} & 88.99 &	90.62	& 89.62 \\ \cline{2-5} 
 & \textit{Recall} & 78.44 &	95.17 &	89.58 \\ \cline{2-5} 
 & \textit{F1-Score} & 83.15 &	92.81 &	89.58 \\ \hline
\end{tabular}
\caption{Experimental results of the proposed visual sentiment analyzer on the task 1 in terms of accuracy, precision, recall and F1-score per class. \textit{P represents the version of the model pre-trained on places dataset while the rest are pre-trained on ImageNet dataset.}}
\label{results_per_class_task1}
\end{table}


Table~\ref{results_per_class_task2} provides the experimental results in terms of accuracy, precision, recall, and F1-score per class of task 2, where seven different categories of sentiments are considered. Overall better results are observed on class \textit{sadness} while lowest performance has been observed on class \textit{anger}, where precision and recall are generally on the lower side for most of the models. 
\begin{table}[!ht]
\centering
\scalebox{0.7}{
\begin{tabular}{|c|c|c|c|c|c|c|c|c|}
\hline
\textbf{Model} & \textbf{Metric} & \textbf{Joy} & \textbf{Sadness} & \textbf{Fear} & \textbf{Diguest} & \textbf{Anger} & \textbf{Surprise} & \textbf{Neutral} \\ \hline
\multirow{4}{*}{\textbf{VGGNet}} & \textit{Accuracy} & 83.37 &	95.32 &	88.24 &	76.67 &	76.86	 & 75.29 &	75.78 \\ \cline{2-9}
& \textit{Precision} & 92.17 &	92.46 &	85.09 &	76.78 &	82.13 &	76.96 &	80.31 \\ \cline{2-9} 
& \textit{Recall} &  76.78 &	99.12 &	94.83 &	60.63 &	56.71 &	77.22 &	73.32 \\ \cline{2-9} 
& \textit{F1-Score}   & 83.77 &	95.67 &	89.68 &	67.68 &	66.99 &	77.07 &	76.35 \\ \hline
\multirow{4}{*}{\textbf{VGGNet (p)}} & \textit{Accuracy} &  84.59 &	95.67 &	88.86 &	76.07 &	77.43 &	75.99 &	77.21  \\ \cline{2-9} 
 & \textit{Precision}  &  92.44 &	93.47 &	85.47 &	71.19 &	76.23 &	75.21 &	81.58 \\ \cline{2-9} 
& \textit{Recall}   &  78.65 &	98.60 &	95.46 &	67.15 &	62.18 &	82.27 &	75.64 \\ \cline{2-9} 
& \textit{F1-Score}   &  84.99 &	95.97 &	90.19 &	68.78 &	68.33 &	78.58 &	78.43 \\ \hline
 \multirow{4}{*}{\textbf{Inception V-3}} & \textit{Accuracy} &  79.81 &	90.51 &	85.40 &	76.26 &	75.51 &	76.21 &	75.51 \\ \cline{2-9} 
 & \textit{Precision} & 89.81 &	86.77 &	81.19 &	86.36 &	86.12 &	71.72 &	75.66 \\ \cline{2-9} 
& \textit{Recall}   &  72.09 &	96.53 &	94.88 &	49.30 &	49.87 &	92.06 &	80.48 \\ \cline{2-9} 
& \textit{F1-Score}   &  79.94 &	91.39 &	87.50 &	62.57 &	62.64 &	80.62 &	77.84 \\ \hline
 \multirow{4}{*}{\textbf{ResNet-50}} & \textit{Accuracy} &  85.59 &	95.03 &	87.97 &	75.16 &	77.64 &	73.75 &	75.72  \\ \cline{2-9} 
& \textit{Precision}  &  94.16 &	92.71 &	86.18 &	73.83 &	81.89 &	79.31 &	78.49  \\ \cline{2-9} 
& \textit{Recall}  &  79.15 &	98.19 &	92.52 &	61.23 &	59.70 &	69.63 &	75.59 \\ \cline{2-9} 
& \textit{F1-Score}   &  85.99 &	95.37 &	89.22 &	66.43 &	68.83 &	73.92 &	76.91 \\ \hline
 \multirow{4}{*}{\textbf{ResNet-101}} & Accuracy &  85.10 &	95.30 &	88.38 &	76.13 &	76.10 &	75.43 &	77.24  \\ \cline{2-9} 
 & \textit{Precision} &  88.15 &	93.59 &	86.84 &	76.67 &	76.90 &	74.76 &	79.28 \\ \cline{2-9} 
& \textit{Recall}   &  84.86 &	97.67 &	92.42 &	58.95 &	61.13 &	82.00 &	77.96 \\ \cline{2-9} 
& \textit{F1-Score}   &  86.42 &	95.59 &	89.54 &	66.49 &	67.94 &	78.17 &	78.60 \\ \hline
 \multirow{4}{*}{\textbf{DenseNet}} & \textit{Accuracy} &  83.81 &	93.51 &	87.32 &	76.24 &	76.48 &	75.78 &	75.43 \\ \cline{2-9} 
& \textit{Precision}  &  91.41 &	91.79 &	85.50 &	81.24 &	87.74 &	73.15 &	77.47 \\ \cline{2-9} 
& \textit{Recall}   &  78.47 &	96.16 &	92.07 &	53.72 &	50.52 &	86.83 &	76.85 \\ \cline{2-9} 
& \textit{F1-Score}   & 84.41 &	93.92 &	88.66 &	64.52 &	64.02 &	79.38 &	76.94 \\ \hline
  \multirow{4}{*}{\textbf{EfficientNet}} & \textit{Accuracy} &  84.40 &	94.84 &	88.38 &	75.70 &	75.78 &	74.83 &	75.67 \\ \cline{2-9} 
 & \textit{Precision} &  91.44 &	93.04 &	86.16 &	75.49 &	78.24 &	74.24 &	80.47 \\ \cline{2-9} 
& \textit{Recall}   &  79.73 &	97.41 &	93.52 &	63.65 &	59.57 &	81.50 &	72.67 \\ \cline{2-9} 
& \textit{F1-Score}   &  85.09 &	95.16 &	89.63 &	67.57 &	66.82 &	77.65 &	76.13 \\ \hline
  \multirow{4}{*}{\textbf{VGGNet (P+I)}} & \textit{Accuracy} &  83.09 &	95.62 &	89.11 &	77.05 &	77.72 &	77.18 &	77.53 \\ \cline{2-9} 
 & \textit{Precision}  &  95.89 &	93.30 &	84.66 &	73.57 &	76.36 &	74.31 &	82.91 \\ \cline{2-9} 
& \textit{Recall}  &  72.66 &	98.71 &	97.33 &	65.50 &	63.15 &	87.78 &	74.46 \\ \cline{2-9} 
& \textit{F1-Score}   &  82.65 & 	95.93  &	90.55 &	69.24 &	68.91 &	80.45 &	78.41 \\ \hline
\end{tabular}}
\caption{Experimental results of the proposed visual sentiment analyzer on the task 2 in terms of accuracy, precision, recall and F1-score per class. \textit{P represents the version of the model pre-trained on places dataset while the rest are pre-trained on ImageNet dataset.}}
\label{results_per_class_task2}
\end{table}

In line with the previous scenario, in Table \ref{results_per_class_task3} provides the results in terms of accuracy, precision, and recall per class. 

\begin{table*}[!ht]
\centering
\begin{tabular}{|c|c|c|c|c|c|c|c|c|c|c|c|}
\hline
\textbf{Model} & \textbf{Metric} & \textbf{Anger} & \textbf{Anxiety} & \textbf{Craving} & \textbf{Pain} & \textbf{Fear} & \textbf{Horror} & \textbf{Joy} & \textbf{Relief} & \textbf{Sadness} & \textbf{Surprise} \\ \hline
\multirow{4}{*}{VGGNet} & Accuracy &  73.87 &	86.16 &	80.73 &	82.29 &	87.52 &	79.12 &	84.21 &	81.23 &	95.22 &	70.70  \\ \cline{2-12} 
& \textit{Precision} &  63.52 &	82.04 &	61.14 &	76.15 &	81.46 &	67.87 &	95.11 &	92.04 &	92.28 &	79.50 \\ \cline{2-12} 
 & \textit{Recall} &  80.28 &	95.39 &	28.40 &	93.73 &	97.88 &	83.85 &	75.09 &	71.63 &	99.59 &	68.24  \\ \cline{2-12} 
 & \textit{F1-Score} &  70.83 &	88.20 &	38.65 &	83.99 &	88.91 &	75.00 &	83.88 &	80.56 &	95.80 &	72.60  \\ \hline
\multirow{4}{*}{VGGNet (p)} & Accuracy & 74.81 &	83.76 &	79.57 &	80.34 &	86.38 &	78.23 &	82.12 &	77.31 &	95.16 &	72.81 \\ \cline{2-12} 
& \textit{Precision} &  64.38 &	77.37 &	60.14 &	73.74 &	80.73 &	69.92 &	95.22 &	93.12 &	92.37 &	75.15  \\ \cline{2-12} 
 & \textit{Recall} &  82.81 &	97.11 &	29.77 &	92.17 &	96.83 &	77.25 &	72.07 &	65.11 &	99.39 &	78.35  \\ \cline{2-12} 
 & \textit{F1-Score} &  72.39 &	86.12 &	39.52 &	81.90 &	88.04 &	73.37 &	82.00 &	76.63 &	95.75 &	76.61  \\ \hline
 \multirow{4}{*}{Inception V-3} & \textit{Accuracy} &  75.76 &	85.29 &	81.40 &	80.96 &	86.24 &	75.70 &	82.43 &	79.46 &	94.36 &	73.03 \\ \cline{2-12} 
& \textit{Precision} &  63.38 &	80.14 &	91.79 &	73.47 &	80.12 &	62.47 &	94.86 &	92.68 &	91.84 &	73.18 \\ \cline{2-12} 
 & \textit{Recall} &  92.00 &	96.93 &	14.66 &	96.47 &	97.23 &	87.80 &	71.99 &	67.74 &	98.42 &	84.89 \\ \cline{2-12} 
 & \textit{F1-Score} &  75.04 &	87.73 &	25.24 &	83.41 &	87.83 &	72.95 &	81.79 &	78.18 &	95.02 &	78.47  \\ \hline
 \multirow{4}{*}{ResNet-50} & \textit{Accuracy} &  72.81 &	85.38 &	79.93 &	81.21 &	86.79 &	79.12 &	85.10 &	81.79 &	94.72 &	71.48  \\ \cline{2-12} 
& \textit{Precision} &  63.91 &	82.12 &	55.65 &	76.23 &	82.89 &	69.13 &	90.09 &	89.59 &	92.78 &	75.17  \\ \cline{2-12} 
 & \textit{Recall} &  71.93 &	93.39 &	32.81 &	90.31 &	93.53 &	79.91 &	81.94 &	75.27 &	97.97 &	76.37  \\ \cline{2-12} 
 & \textit{F1-Score} &  67.41 &	87.39 &	41.13 &	82.67 &	87.87 &	74.08 &	85.77 &	81.78 &	95.30 &	75.59 \\ \hline
 \multirow{4}{*}{ResNet-101} & \textit{Accuracy} &  73.09 &	85.40 &	79.84 &	82.71 &	87.46 &	78.62 &	85.29 &	80.87 &	94.72 &	72.31 \\ \cline{2-12} 
 & \textit{Precision} &  63.08 &	81.02 &	55.52 &	76.32 &	83.32 &	70.46 &	93.01 &	90.90 &	92.54 &	76.84  \\ \cline{2-12} 
 & \textit{Recall} &  77.40 &	95.49 &	32.30 &	94.51 &	94.40 &	74.11 &	79.30 &	72.04 &	98.27 &	75.08  \\ \cline{2-12} 
 & \textit{F1-Score} &  69.49 &	87.66 &	40.72 &	84.44 &	88.50 &	72.14 &	85.52 &	80.36 &	95.32 &	75.90 \\ \hline
 \multirow{4}{*}{DenseNet} & \textit{Accuracy} & 73.31 &	84.88 &	80.73 &	81.10 &	87.21 &	77.95 &	82.60 &	81.26 &	93.41 &	72.17  \\ \cline{2-12} 
 & \textit{Precision} & 63.80 &	81.41 &	67.75 &	74.75 &	83.03 &	66.11 &	90.24 &	89.92 &	92.33 &	74.64  \\ \cline{2-12} 
 & \textit{Recall} & 75.51 &	93.50 &	19.33 &	93.50 &	94.29 &	84.60 &	76.76 &	73.84 &	95.93 &	79.10 \\ \cline{2-12} 
 & \textit{F1-Score} & 67.20&	87.92&	38.81&	84.44&	88.22&	75.20&	83.16&	80.12&	95.37&	77.37 \\ \hline
 \multirow{4}{*}{EfficientNet} &  \textit{Accuracy} &  74.46&	86.34&	81.40&	83.18&	88.12&	77.62&	82.72&	78.24&	94.91&	72.38  \\ \cline{2-12} 
 & \textit{Precision} &  62.07&	82.16&	65.22&	76.69&	84.58&	71.05&	91.86&	91.26&	92.43&	79.86  \\ \cline{2-12} 
 & \textit{Recall} &  79.08&	95.43&	31.80&	95.71&	94.55&	70.48&	75.96&	68.16&	98.56&	65.82 \\ \cline{2-12} 
 & \textit{F1-Score} &  69.06 &	87.03 &	30.02 &	83.07 &	88.28 &	74.09 &	82.85 &	81.05 &	94.08 &	76.72 \\ \hline
  \multirow{4}{*}{VGGNet (P+I)} &  \textit{Accuracy} &  75.90 &	84.24 &	79.96 &	80.43 &	87.24 &	78.23 &	82.35 &	78.32 &	95.47 &	73.56 \\ \cline{2-12} 
 & \textit{Precision} &  64.85 &	77.44 &	66.01 &	72.59 &	81.06 &	67.11 &	95.87 &	94.75 &	92.61 &	77.24  \\ \cline{2-12} 
 & \textit{Recall} &  86.71 &	98.23 &	24.85 &	95.57 &	98.34 &	86.27 &	71.93 &	65.69 &	99.70 &	76.10 \\ \cline{2-12} 
 & \textit{F1-Score} &  74.07 &	86.60 &	35.68 &	82.50 &	88.87 &	75.49 &	82.16 &	77.59 &	96.02 &	76.55 \\ \hline
\end{tabular}
\caption{Experimental results of the proposed visual sentiment analyzer on the task 3 in terms of accuracy, precision, recall and F1-score per class.}
\label{results_per_class_task3}
\end{table*}

\subsection{Lessons learned}
This initial work on visual sentiment analysis has revealed a number of challenges, showing us all the different facets of such a complex research domain. We have summarized the main points hereafter:
\begin{itemize}
    \item Sentiment analysis aims to extract people's perceptions of the images; thus, crowd-sourcing seems a suitable option for collecting training and ground truth datasets. However, choosing labels/tags for conducting a successful crowd-sourcing study is not straightforward. 
    \item The most commonly used three tags, namely positive, negative, and neutral are not enough to fully exploit the potential of visual sentiment analysis in applications like disaster analysis. The complexity of the task increases as we go deeper into sentiment/emotion hierarchy.
    
    \item The majority of the disaster-related images in social media represent negative (i.e., sad, horror, pain, anger, and fear etc.,) sentiments; however, we noticed that there exists a number of samples able to evoke positive emotions, such as joy and relief. 
    
    \item  Disaster-related images from social media exhibit sufficient features to evoke human emotions. Object in images (gadget, clothes, broken houses, scene-level (i.e., background, landmarks), color/contrast, and human expressions, gestures, and poses provide crucial cues in the visual sentiment analysis of disaster-related images. This can be a valuable aspect to be considered to represent people's emotions and sentiments.
    
    
    \item Human emotions and sentiments tags are correlated, as can also be noticed from the statistics of the crowd-sourcing study, thus a multi-label framework is likely to be most promising research direction.

\end{itemize}

\section{Conclusions and Future research directions}
\label{conclusion}
In this article, we focused on the emerging concept of visual sentiment analysis, and showed how natural disaster-related images evoke people's emotions and sentiments. To this aim, we proposed a pipeline starting from data collection and annotation via a crowd-sourcing study, and conclude with the development and training of deep learning models for multi-label classification of sentiments. In the crowd-sourcing study, we analyzed and annotated 4,003 images with four different set of tags resulting in four different datasets with different hierarchies of sentiments. Based on our analysis, we believe visual sentiment analysis in general and the analysis of natural disaster-related content, in particular, is an exciting research domain that will benefit users and the community in a diversified set of applications. 
The current literature shows a tendency towards visual sentiment analysis of general images shared on social media by deploying deep learning techniques to extract object and facial expression based visual cues. However, we believe, as also demonstrated in this work, visual sentiment analysis can be extended to more complex images where several types of image features and information, such as object and scene-level features, human expressions, gestures and poses, could be jointly utilized. All this can be helpful to introduce new applications and services. 

We believe, there is a lot to be explored yet in this direction, and this work provides a baseline for future work in the domain. In the future, we would like to collect a multi-model dataset where the text associated with images complements visual features leading to improved visual sentiment analysis. 

\bibliographystyle{unsrt}
\bibliography{References}

\end{document}